\documentclass{article}

\usepackage{microtype}
\usepackage{graphicx}
\usepackage{subfigure}
\usepackage{booktabs} 
\usepackage{tabularx}
\newcolumntype{L}[1]{>{\raggedright\let\newline\\\arraybackslash\hspace{0pt}}m{#1}}
\newcolumntype{C}[1]{>{\centering\let\newline\\\arraybackslash\hspace{0pt}}m{#1}}
\newcolumntype{R}[1]{>{\raggedleft\let\newline\\\arraybackslash\hspace{0pt}}m{#1}}
\usepackage{amsmath}
\usepackage[hyphens]{url}

\usepackage{hyperref}

\usepackage[accepted]{icml2019}

\icmltitlerunning{On improving deep learning generalization with adaptive sparse connectivity}

\begin{document}

\twocolumn[
\icmltitle{On improving deep learning generalization with adaptive sparse connectivity}




\begin{icmlauthorlist}
\icmlauthor{Shiwei Liu}{to}
\icmlauthor{Decebal Constantin Mocanu}{to}
\icmlauthor{Mykola Pechenizkiy}{to}

\end{icmlauthorlist}

\icmlaffiliation{to}{Department of Mathematics and Computer Science, Eindhoven University of Technology, Eindhoven, Netherlands}

\icmlcorrespondingauthor{Shiwei Liu}{s.liu3@tue.nl}

\icmlkeywords{Machine Learning, ICML}

\vskip 0.3in
]



\printAffiliationsAndNotice{}  

\begin{abstract}
Large neural networks are very successful in various tasks. However, with limited data, the generalization capabilities of deep neural networks are also very limited. In this paper, we empirically start showing that intrinsically sparse neural networks with adaptive sparse connectivity, which by design have a strict parameter budget during the training phase, have better generalization capabilities than their fully-connected counterparts. Besides this, we propose a new technique to train these sparse models by combining the Sparse Evolutionary Training (SET) procedure with neurons pruning. Operated on MultiLayer Perceptron (MLP) and tested on 15 datasets, our proposed technique zeros out around 50\% of the hidden neurons during training, while having a linear number of parameters to optimize with respect to the number of neurons. The results show a competitive classification and generalization performance. 
\end{abstract}

\section{Introduction}
\label{submission}

In spite of the good performance of deep neural networks, they encounter generalization issues and overfitting problems, especially when the amount of parameters is much higher than the amount of training examples. While, understanding this trade-off is an open research question, fortunately, various works have been proposed to handle this problem including, implicit norm regularization \cite{neyshabur2014search}, Two-Stage Training Process \cite{zheng2018improvement}, dropout \cite{srivastava2014dropout}, Batch normalization \cite{ioffe2015batch}, etc. Recently, many complexity measures have emerged to understand what drives generalization in deep networks, such as sharpness \cite{keskar2016large}, PAC-Bayes \cite{dziugaite2017computing} and margin-based measures \cite{neyshabur2017pac}. \cite{neyshabur2017exploring} analyze different complexity measures and demonstrate that the combination of some of these measures seems to capture better neural networks generalization behavior. 

On the other side, the ability of sparse neural networks to reduce the number of parameters can dramatically shrink the model size and, therefore, relieve overfitting. However, the traditional algorithms to train such networks, make use of an initial fully-connected network which is trained first. Further on, the unimportant connections in this network are pruned using various techniques, e.g. \cite{lecun1990optimal, hassibi1993second, han2017ese, narang2017exploring, lee2018snip} to obtain a sparse topology. The initial fully-connected network is a critical point hindering neural networks scalability due to its quadratic number of (many unnecessary) parameters with respect to its number of neurons.  To address this issue, \cite{mocanu2018scalable} have proposed a new class of models, i.e.  intrinsically sparse neural networks with adaptive sparse connectivity. These models have a linear number of parameters with respect to the number of neurons, don't require an initial fully-connected network, and can be trained with the Sparse Evolutionary Training (SET) procedure.

In this paper, we introduce a new improvement to SET, dubbed SET with Neurons Pruning (NPSET), to further reduce the number of hidden neurons and parameters. Our approach is able to identify and eliminate a large number of non-informative hidden neurons and their accompanying connections by applying neurons pruning to the SET procedure. Same as SET, NPSET starts with a sparse topology, thus having a clear advantage over the state-of-the-art methods which start from fully connected topologies. The experimental results show that the removal of the hidden layer neurons with very few output connections allows NPSET to further reduce computational costs in both phases (training and inference). Moreover, we show that intrinsically sparse MLPs trained with both, SET or NPSET, have higher generalization ability than their fully-connected counterparts.

\section{Related Work}
\begin{table}[h!]
\scriptsize
\caption{\label{tabledata}Datasets characteristics.}
\begin{center}
\setlength\tabcolsep{1.5pt}
\begin{tabular}{ |l| l |l |l |l |l |l|}
\hline
\textbf{Dataset} & \textbf{Number of} & \textbf{Features} & \textbf{Data Type} & \textbf{Classes} & \textbf{Training}  & \textbf{Test} \\
&\textbf{Samples} & &  & & \textbf{Samples}  & \textbf{Samples} \\
\hline
Leukemia & 72 & 7070 & Discrete & 2 & 48  & 24 \\
\hline
PCMAC & 1943 & 3289 & Discrete & 2 & 1295  & 648 \\
\hline
Lung-discrete & 73 & 325 & Discrete & 7 & 48  & 25 \\
\hline
gisette & 7000 & 5000 & Continuous & 2 & 4666 & 2334 \\
\hline
lung & 203 & 3312 &Ccontinuous & 5 & 135  & 68 \\
\hline
CLL-SUB-111 & 111 & 11340 & Continuous & 3 & 74  & 37 \\
\hline
Carcinom & 174 & 9183 & Continuous & 11 & 116 & 58 \\
\hline
orlraws10P & 100 & 10304 & Continuous & 10 & 66  & 34 \\
\hline
TOX-171 & 171 & 5748 & Continuous & 4 & 114  & 57 \\
\hline
Prostate-GE & 102 & 5966 & Continuous & 2 & 68  & 34 \\
\hline
arcene & 200 & 10000 & Continuous & 2 & 133  & 67 \\
\hline
madelon & 2600 & 500 & Continuous & 2 & 1733  & 867 \\
\hline
Yale & 165 & 1024 & Continuous & 15 & 110  & 55 \\
\hline
GLIOMA & 50 & 4434 & Continuous & 4 & 33  & 17 \\
\hline
RELATHE & 1427 & 4322 & Continuous & 2 & 951  & 476 \\
\hline
\end{tabular}
\end{center}
\vskip -0.15in
\end{table}
Inspired by Darwinian theory, Sparse Evolutionary Training (SET) \cite{mocanu2018scalable} is a simple but efficient training method which enables an initially sparse topology of bipartite layers of neurons to evolve towards a scale-free topology, while learning to fit the data characteristics. After each training epoch, the connections having weights closest to zero are removed (magnitude based removal). After that, new connections (in the same  amount as the removed ones) are randomly added to the network. This offers benefits in both, computational time (pronounced faster training and testing time in comparison with fully connected bipartite layers) and quadratically lower memory requirements. The interested reader is referred to \cite{mocanu2018scalable} for a detailed discussion, and to \cite{dynamicreparaterimzation, federatedlearning, lowmemory} for further developments and analyses on it.
\begin{table*}[h]
\footnotesize
\vskip -0.1in
\caption{\label{table3}The maximum accuracy of each method for each dataset. The entry with the highest accuracy for each dataset is made bold. }
\begin{center}
\vskip 0in
\begin{tabular}{ |l| l| l| l| l| l| l|}
\hline
\textbf{Dataset} & \textbf{SET-} & \textbf{NPSET-} & \textbf{$1^{st}$ NPSET-} & \textbf{$2^{nd}$ NPSET-} & \textbf{Direct} & \textbf{Direct}\\
 & \textbf{MLP (\%)} & \textbf{MLP (\%)} & \textbf{MLP (\%)} & \textbf{MLP (\%)} & \textbf{SET-MLP (\%)} & \textbf{FC-MLP (\%)}\\
\hline
Leukemia & \textbf{87.50}  & \textbf{87.50 (+0.00)} & \textbf{87.50 (+0.00)} & \textbf{87.50 (+0.00)} & \textbf{87.50 (+0.00)} & 75.00 (-12.50) \\
\hline
PCMAC & 87.35 & \textbf{88.43 (+1.08)} & 86.73 (-0.62) & \textbf{88.43 (+1.08)} & 87.81 (+0.46) & 85.19 (-2.16) \\
\hline
Lung-discrete & \textbf{88.00} & \textbf{88.00 (+0.00)} & \textbf{88.00 (+0.00)} & 84.00 (-4.00) & \textbf{88.00 (+0.00)} & 80.00 (-8.00)  \\
\hline
gisette & 97.43  & 97.52 (+0.09) & \textbf{97.64 (+0.21)} & 97.52 (+0.09) & 97.47 (+0.04) & 97.60 (+0.17) \\
\hline
lung & 92.65 & \textbf{94.12 (+1.47)}  & \textbf{94.12 (+1.47)} & 92.65 (+0.00) & \textbf{94.12 (+1.47)} & 92.65 (+0.00) \\
\hline
CLL-SUB-111 & 67.57 & \textbf{75.68 (+8.11)} & 62.16 (-5.41) & 67.57 (+0.00) & 70.27 (+2.70) & 59.46 (-8.11) \\
\hline
Carcinom & 79.31 & \textbf{81.03 (+1.72)} & 75.86 (-3.45) & 75.86 (-3.45) & 77.59 (-1.72) & 68.97 (-10.34) \\
\hline
orlraws10P & \textbf{88.24} & \textbf{88.24 (+0.00)} & 85.29 (-2.95) & \textbf{88.24 (+0.00)} & \textbf{88.24 (+0.00)} & 79.41 (-8.77) \\
\hline
TOX-171 & \textbf{91.23} & \textbf{91.23 (+0.00)} & 85.97 (-5.26) & 89.47 (-1.76) & \textbf{{91.23} (+0.00)} & 82.46 (-8.77) \\
\hline
Prostate-GE & \textbf{88.24} & \textbf{88.24 (+0.00)} & \textbf{88.24 (+0.00)} & \textbf{88.24 (+0.00)} & \textbf{88.24 (+0.00)} & 79.41 (-8.83) \\
\hline
arcene & 77.61 & 77.61 (+0.00) & \textbf{82.09 (+4.48)} & 74.63 (-2.98) & 79.10 (+1.49) & 77.61 (+0.00) \\
\hline
madelon & 71.16 & 71.28 (+0.12) & \textbf{71.74 (+0.58)} & 70.13 (-1.03) & 71.05 (-0.11) & 56.40 (-14.76) \\ 
\hline
Yale & 70.91 & \textbf{74.55 (+3.64)} & 69.09 (-1.82) & 70.91 (+0.00) & 69.09 (-1.82) & 63.64 (-7.27) \\
\hline
GLIOMA & \textbf{76.47} & \textbf{76.47 (+0.00)} & \textbf{76.47 (+0.00)} & \textbf{76.47 (+0.00)} & \textbf{76.47 (+0.00)} & 64.71 (-11.76) \\
\hline
RELATHE & 89.71 & 90.55 (+0.84) & 89.71 (+0.00) & 89.92 (+0.21) & 87.61 (-2.10) & \textbf{90.76 (+1.05)}\\
\hline

\end{tabular}
\end{center}
\vskip -0.2in
\end{table*}
\section{Methods}
In this section, we detail our proposed method (NPSET).
\begin {figure}
\vskip -0.1in
\centerline{\includegraphics[width=0.9\columnwidth]{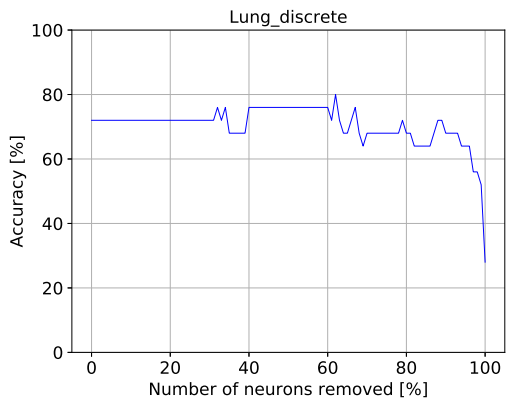}}
\caption{\label{neuronsremoved}Influence of hidden neurons removal (from the first hidden layer) on accuracy on the Lung-discrete dataset.}
\vskip -0.1in
\end{figure}
\subsection{Why Neurons Pruning.}
The sparse topology allows SET to create MultiLayer Perceptrons with hundreds of thousands of neurons \cite{liu2019sparse}, which guarantees their ability to represent all sorts of features and approximate function to tackle different problems. However, such a large number of neurons is also a double-edged sword which can lead to significant redundancy. For example, in the case of CIFAR10 dataset whose features number is 3072, the neurons number of the first hidden layer is 4000 in \cite{mocanu2018scalable}. Obviously, not all neurons can provide important information to outputs. To prove this, we test whether removing hidden neurons which have the least output connections will decrease the performance. Figure \ref{neuronsremoved} shows the influence of removing neurons from the first hidden layer for Lung-discrete dataset (due to the space limitation, we only give one dataset). It can be observed that the model maintains or even improves its accuracy after removing these unimportant neurons. In order to remove these non-informative neurons, at the beginning of each training epoch, we remove a certain fraction $\alpha$ of hidden neurons that have the smallest numbers of connections. Thus, with a large probability they will not have a notable impact on the model performance. 
\begin{table*}[h]
\footnotesize
\caption{\label{table4}Compression Rates of SET-MLP and NPSET-MLP based on the detailed below FC-MLP.}
\begin{center}

\begin{tabular}{|l| r| r| r| r| r| r| R{1.5cm}| r|}
\hline
\textbf{Dataset} & \multicolumn{3}{c|}{\textbf{Parameters (\#)}} & \multicolumn{3}{c|}{\textbf{Neurons (\#)}} & \multicolumn{2}{c|}{\textbf{Compression Rate ($\times$)}}\\  \cline{2-4}  \cline{5-7}  \cline{8-9} 
& \tiny{\textbf{FC-MLP}} & \tiny{\textbf{SET-MLP}} & \tiny{\textbf{NPSET-MLP}} & \tiny{\textbf{FC-MLP}} &\tiny{\textbf{SET-MLP}}& \tiny{\textbf{NPSET-MLP}} & \tiny{\textbf{SET-MLP}} & \tiny{\textbf{NPSET-MLP}} \\
\hline
Leukemia & 98,504,000 & 294,235 & 40,039 & 21,070 & 21,070 & 9,710 & 335$\times$ & 2,460$\times$\\
\hline
PCMAC & 18,873,000 & 128,432 & 18,622 & 9,289  & 9,289 & 4,435 & 147$\times$ & 1,013$\times$ \\
\hline
Lung-discrete & 189,600 & 13,446 & 2,447 & 925 & 925 & 457 & 14$\times$ & 77$\times$\\
\hline
gisette & 50,010,000 & 209,556 & 29,884 & 15,000 & 15,000 & 6,892 & 238$\times$ & 1,673$\times$\\
\hline
lung & 18,951,000 & 135,689 & 19,776 & 9,312 & 9,312& 4,458 &140$\times$ & 958$\times$\\
\hline
CLL-SUB-111 & 245,773,000 & 474,738 & 65,421  & 33,340& 33,340 & 15,488 &518$\times$ & 3,757$\times$\\
\hline
Carcinom & 163,746,000 & 420,592 & 67,726 & 27,182 & 27,182& 12,580 &389$\times$ & 2,418$\times$\\
\hline
orlraws10P & 203,140,000 & 465,977 & 72,871 & 30,304& 30,304& 14,072& 436$\times$ & 2,788$\times$ \\
\hline
TOX-171 & 53,760,000 & 225,416 & 31,815  & 15,748& 15,748 & 7,640& 238$\times$ & 1,690$\times$ \\
\hline
Prostate-GE & 54,840,000 & 219,191 & 30,690 & 15,966& 15,966 & 7,858 & 250$\times$ & 1,787$\times$ \\
\hline
arcene & 200,020,000 & 419,469 & 57,136 & 30,000 & 30,000 & 13,768 &477$\times$ & 3,501$\times$ \\
\hline
madelon & 1,502,000 & 36,563 & 5,096 & 2,500  & 2,500 & 896 & 41$\times$ & 295$\times$ \\
\hline
Yale & 2,039,000 & 47,222 & 8,576 & 3,024 & 3,024 & 1,420  &43$\times$ & 238$\times$ \\
\hline
GLIOMA & 33,752,000 & 178,678 & 25,228 & 12,434 & 12,434 & 5,956  &189$\times$ & 1,338$\times$ \\
\hline
RELATHE & 33,296,000 & 170,804 & 24,280 & 12,322& 12,322 & 5,844  &195$\times$ & 1,371$\times$ \\
\hline

\end{tabular}
\end{center}
\vskip -0.15in
\end{table*}
\subsection{Where to Start Pruning.}
The initial network topology generated by SET is randomly sparse and does not provide any specific information. Thus, pruning neurons at the beginning may eliminate significant neurons forever, along with a serious damage to performance. It is best to start applying neurons pruning after a certain number $\beta$ of epochs rather than removing neurons from the beginning. After evolving in the first $\beta$ epochs, the network has already learned how to identify and retain important connections, while the evolved neurons connectivity provides a helpful guidance to identify non-important neurons.

\subsection{How Many Epochs to Prune.}
It is the fact that if we prune neurons in each epoch, the final number of neurons would be too small to keep a good accuracy. On the other hand, if we only prune neurons for several epochs, the number of removed neurons would be too trivial to decrease the computation. To preserve the good performance, we only apply neurons pruning for $\gamma$ epochs, after which the SET procedure continues normally.

\section{Experiments and Results}

We evaluated the proposed NPSET\footnote{The code of NPSET is built based on the source code of SET \url{ https://github.com/dcmocanu/sparse-evolutionary-artificial-neural-networks}.} method by training sparse MLPs from scratch on 15 classification datasets with limited amount of samples and many input features, as detailed in Table~\ref{tabledata}. All datasets can be retrieved from Arizona State University open-source repository\footnote{\url{http://featureselection.asu.edu/index.php}.}.
In order to understand the NPSET performance better, we compare it against five methods: (1) SET-MLP \cite{mocanu2018scalable}, (2) $1^{st}$ NPSET-MLP where only neurons of the first hidden layer are pruned, (3) $2^{nd}$ NPSET-MLP where only neurons of the second hidden layer are pruned, 
(4) Direct SET-MLP, directly trained SET-MLP having the same hidden layer sizes as NPSET-MLP after neurons pruning. (5) Direct FC-MLP, directly trained FC-MLP having the same hidden layer sizes as NPSET-MLP after neurons pruning. All models used in this paper have two hidden layers, and ReLU activation function. We trained NPSET-MLP on a python implementation of fully connected MLPs\footnote{\url{https://github.com/ritchie46/vanilla-machine-learning}.}, as the SET-MLP implementation was also built on top of this code, guaranteeing the validity of the comparison in this paper. Since our new method is an improvement over SET, we used SET-MLP as the baseline for experiments. All these methods are trained from scratch. 

To find the most suitable hyperparameters values, we performed a small random search experiment. This showed $\alpha = 0.04$, $\beta = 10$ and $\gamma = 40$ are safe choices that not only removed the non-informative neurons, but also lead NPSET-MLP to have better performance.

The maximum accuracies of all 6 models for each dataset are shown in Table \ref{table3}. From Table \ref{table3}, we can observe that, compared with SET-MLP, NPSET-MLP improves the peak accuracy on 8 datasets, while both models reach better accuracy than their fully-connected counterpart. 
\begin {figure*}[h!]
\vskip -1in
\centerline{\includegraphics[width=170mm,height=160mm]{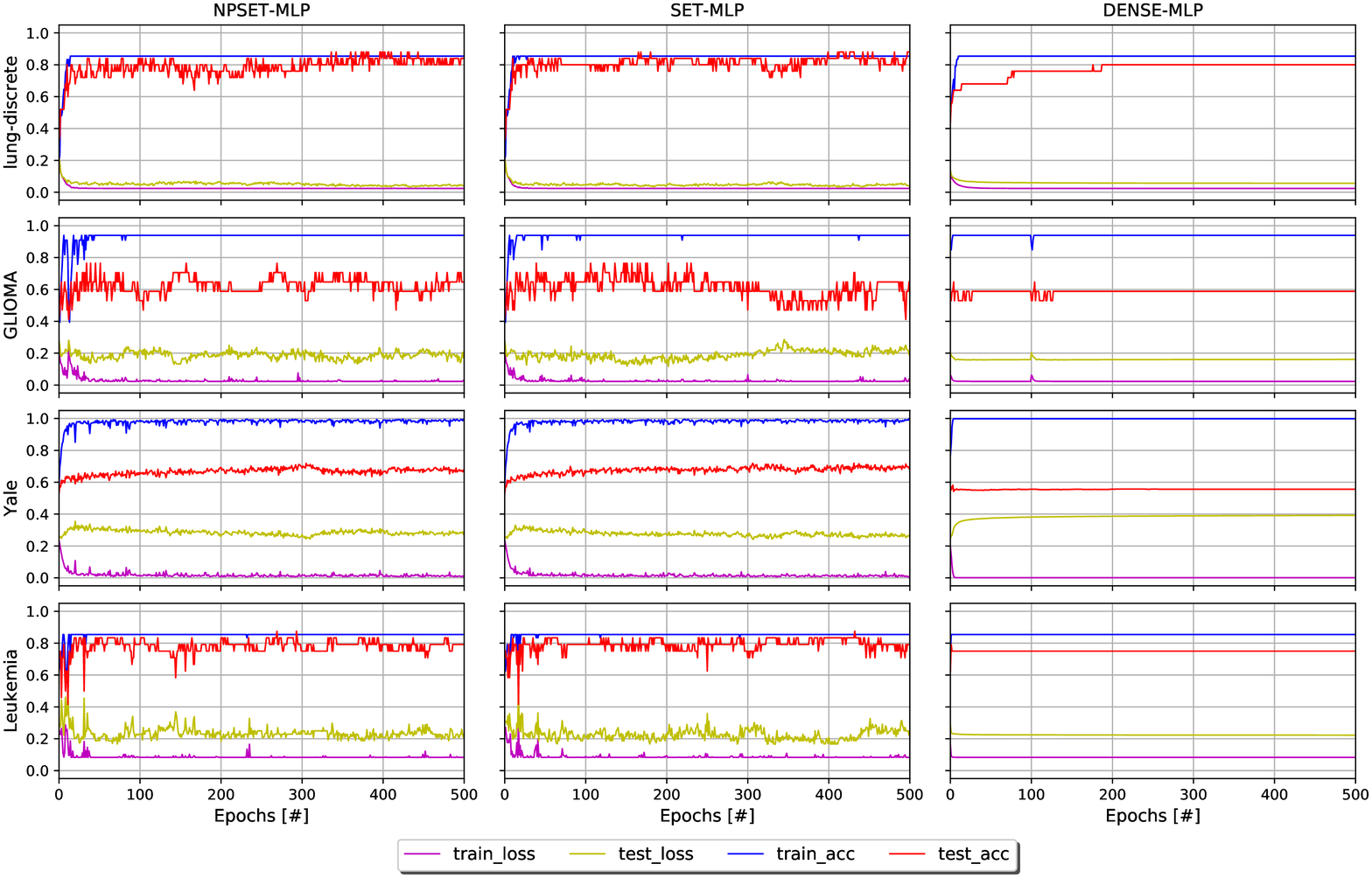}}
\vskip -0.15in
\caption{\label{generalization} NPSET-MLP, SET-MLP and Dense-MLP generalization capabilities reflected by their learning curves.}
\vskip -0.2in
\end{figure*}
Table \ref{table4} shows the compression rates, the numbers of parameters and neurons on 15 datasets for SET-MLP and NPSET-MLP. It worths noting that applying iterative neurons pruning to SET-MLP further increases the compression rate by 6 to 7 times.

To start understanding better the generalization capabilities of SET-MLP and NPSET-MLP, we performed an extra experiment by comparing them with a Dense-MLP (a FC-MLP having the same amount of hidden neurons as SET-MLP). Figure \ref{generalization} shows their learning curves and visualizes their generalization capabilities on 4 datasets (we limited the number of datasets due to space constraints). All three models are trained without any explicit regularization methods e.g. dropout, L1 and L2 Regularization, etc. We can see that the gap between the training and test accuracies of NPSET-MLP and SET-MLP is smaller than for Dense-MLP. Perhaps, the most interesting behavior is on the Yale dataset on which Dense-MLP presents perfect overfitting (e.g. zero loss, 100\% classification accuracy). Contrary, the implicit regularization made by connections addition and removal in SET-MLP and NPSET-MLP don't let these models to perfectly overfit on the training data and enables them better generalization. 

\section{Conclusion}

In this paper we  propose a new method, i.e. NPSET, to enhance the Sparse Training Evolutionary procedure with Neurons Pruning. NPSET trains efficiently intrinsically sparse MLPs, in a number of cases achieving better classification accuracy than SET, while leading to a smaller amount of parameters. This is highly desirable to enhance neural networks scalability. Moreover, the experimental results demonstrate that both methods, SET and NPSET, can train intrinsically sparse MLPs with adaptive sparse connectivity to have higher generalization capabilities than their fully-connected counterparts.

This study is limited in its purpose. E.g., we focus just on MLP which even if can be one of the most used model in real-world applications (it represents 61\% of a typical Google TPU (Tensor Processing Unit) workload~\cite{jouppi2017datacenter}), it does not represent all neural network models. Consequently, there are many future research directions, e.g. analyze the methods performance on much larger tabular datasets, or on other types of neural networks models (e.g. convolutional neural networks). Among all of these, the most interesting research direction would be to understand \textit{why} and \textit{when} intrinsically sparse neural networks with adaptive sparse connectivity can \textit{generalize} better than their fully-connected counterparts.

\bibliography{mybib}

\begin{thebibliography}{19}
\providecommand{\natexlab}[1]{#1}
\providecommand{\url}[1]{\texttt{#1}}
\expandafter\ifx\csname urlstyle\endcsname\relax
  \providecommand{\doi}[1]{doi: #1}\else
  \providecommand{\doi}{doi: \begingroup \urlstyle{rm}\Url}\fi

\bibitem[Dziugaite \& Roy(2017)Dziugaite and Roy]{dziugaite2017computing}
Dziugaite, G.~K. and Roy, D.~M.
\newblock Computing nonvacuous generalization bounds for deep (stochastic)
  neural networks with many more parameters than training data.
\newblock \emph{arXiv preprint arXiv:1703.11008}, 2017.

\bibitem[Han et~al.(2017)Han, Kang, Mao, Hu, Li, Li, Xie, Luo, Yao, Wang,
  et~al.]{han2017ese}
Han, S., Kang, J., Mao, H., Hu, Y., Li, X., Li, Y., Xie, D., Luo, H., Yao, S.,
  Wang, Y., et~al.
\newblock Ese: Efficient speech recognition engine with sparse lstm on fpga.
\newblock In \emph{Proceedings of the 2017 ACM/SIGDA International Symposium on
  Field-Programmable Gate Arrays}, pp.\  75--84. ACM, 2017.

\bibitem[Hassibi \& Stork(1993)Hassibi and Stork]{hassibi1993second}
Hassibi, B. and Stork, D.~G.
\newblock Second order derivatives for network pruning: Optimal brain surgeon.
\newblock In \emph{Advances in neural information processing systems}, pp.\
  164--171, 1993.

\bibitem[Ioffe \& Szegedy(2015)Ioffe and Szegedy]{ioffe2015batch}
Ioffe, S. and Szegedy, C.
\newblock Batch normalization: Accelerating deep network training by reducing
  internal covariate shift.
\newblock \emph{arXiv preprint arXiv:1502.03167}, 2015.

\bibitem[Jouppi et~al.(2017)Jouppi, Young, Patil, Patterson, Agrawal, Bajwa,
  Bates, Bhatia, Boden, Borchers, et~al.]{jouppi2017datacenter}
Jouppi, N.~P., Young, C., Patil, N., Patterson, D., Agrawal, G., Bajwa, R.,
  Bates, S., Bhatia, S., Boden, N., Borchers, A., et~al.
\newblock In-datacenter performance analysis of a tensor processing unit.
\newblock In \emph{Computer Architecture (ISCA), 2017 ACM/IEEE 44th Annual
  International Symposium on}, pp.\  1--12. IEEE, 2017.

\bibitem[Keskar et~al.(2016)Keskar, Mudigere, Nocedal, Smelyanskiy, and
  Tang]{keskar2016large}
Keskar, N.~S., Mudigere, D., Nocedal, J., Smelyanskiy, M., and Tang, P. T.~P.
\newblock On large-batch training for deep learning: Generalization gap and
  sharp minima.
\newblock \emph{arXiv preprint arXiv:1609.04836}, 2016.

\bibitem[LeCun et~al.(1990)LeCun, Denker, and Solla]{lecun1990optimal}
LeCun, Y., Denker, J.~S., and Solla, S.~A.
\newblock Optimal brain damage.
\newblock In \emph{Advances in neural information processing systems}, pp.\
  598--605, 1990.

\bibitem[Lee et~al.(2018)Lee, Ajanthan, and Torr]{lee2018snip}
Lee, N., Ajanthan, T., and Torr, P.~H.
\newblock Snip: Single-shot network pruning based on connection sensitivity.
\newblock \emph{arXiv preprint arXiv:1810.02340}, 2018.

\bibitem[Liu et~al.(2019)Liu, Mocanu, Matavalam, Pei, and
  Pechenizkiy]{liu2019sparse}
Liu, S., Mocanu, D.~C., Matavalam, A. R.~R., Pei, Y., and Pechenizkiy, M.
\newblock Sparse evolutionary deep learning with over one million artificial
  neurons on commodity hardware.
\newblock \emph{arXiv preprint arXiv:1901.09181}, 2019.

\bibitem[Mocanu et~al.(2018)Mocanu, Mocanu, Stone, Nguyen, Gibescu, and
  Liotta]{mocanu2018scalable}
Mocanu, D.~C., Mocanu, E., Stone, P., Nguyen, P.~H., Gibescu, M., and Liotta,
  A.
\newblock Scalable training of artificial neural networks with adaptive sparse
  connectivity inspired by network science.
\newblock \emph{Nature Communications}, 9\penalty0 (1):\penalty0 2383, 2018.

\bibitem[Mostafa \& Wang(2019)Mostafa and Wang]{dynamicreparaterimzation}
Mostafa, H. and Wang, X.
\newblock Parameter efficient training of deep convolutional neural networks by
  dynamic sparse reparameterization.
\newblock \emph{CoRR}, abs/1902.05967, 2019.
\newblock URL \url{http://arxiv.org/abs/1902.05967}.

\bibitem[Narang et~al.(2017)Narang, Elsen, Diamos, and
  Sengupta]{narang2017exploring}
Narang, S., Elsen, E., Diamos, G., and Sengupta, S.
\newblock Exploring sparsity in recurrent neural networks.
\newblock \emph{arXiv preprint arXiv:1704.05119}, 2017.

\bibitem[Neyshabur et~al.(2014)Neyshabur, Tomioka, and
  Srebro]{neyshabur2014search}
Neyshabur, B., Tomioka, R., and Srebro, N.
\newblock In search of the real inductive bias: On the role of implicit
  regularization in deep learning.
\newblock \emph{arXiv preprint arXiv:1412.6614}, 2014.

\bibitem[Neyshabur et~al.(2017{\natexlab{a}})Neyshabur, Bhojanapalli,
  McAllester, and Srebro]{neyshabur2017exploring}
Neyshabur, B., Bhojanapalli, S., McAllester, D., and Srebro, N.
\newblock Exploring generalization in deep learning.
\newblock In \emph{Advances in Neural Information Processing Systems}, pp.\
  5947--5956, 2017{\natexlab{a}}.

\bibitem[Neyshabur et~al.(2017{\natexlab{b}})Neyshabur, Bhojanapalli, and
  Srebro]{neyshabur2017pac}
Neyshabur, B., Bhojanapalli, S., and Srebro, N.
\newblock A pac-bayesian approach to spectrally-normalized margin bounds for
  neural networks.
\newblock \emph{arXiv preprint arXiv:1707.09564}, 2017{\natexlab{b}}.

\bibitem[Sohoni et~al.(2019)Sohoni, Aberger, Leszczynski, Zhang, and
  R{\'e}]{lowmemory}
Sohoni, N.~S., Aberger, C.~R., Leszczynski, M., Zhang, J., and R{\'e}, C.
\newblock Low-memory neural network training: A technical report.
\newblock \emph{CoRR}, abs/1904.10631, 2019.

\bibitem[Srivastava et~al.(2014)Srivastava, Hinton, Krizhevsky, Sutskever, and
  Salakhutdinov]{srivastava2014dropout}
Srivastava, N., Hinton, G., Krizhevsky, A., Sutskever, I., and Salakhutdinov,
  R.
\newblock Dropout: a simple way to prevent neural networks from overfitting.
\newblock \emph{The Journal of Machine Learning Research}, 15\penalty0
  (1):\penalty0 1929--1958, 2014.

\bibitem[Zheng et~al.(2018)Zheng, Yang, Yang, Zhang, and
  Zhang]{zheng2018improvement}
Zheng, Q., Yang, M., Yang, J., Zhang, Q., and Zhang, X.
\newblock Improvement of generalization ability of deep cnn via implicit
  regularization in two-stage training process.
\newblock \emph{IEEE Access}, 6:\penalty0 15844--15869, 2018.

\bibitem[Zhu \& Jin(2018)Zhu and Jin]{federatedlearning}
Zhu, H. and Jin, Y.
\newblock Multi-objective evolutionary federated learning.
\newblock \emph{CoRR}, abs/1812.07478, 2018.

\end{thebibliography}
\bibliographystyle{icml2019}

\end{document}